# Progressive Translation of H&E to IHC with Enhanced Structural Fidelity

Yuhang Kang[a] †, Ziyu Su [a] * †, Tianyang Wang[a], Zaibo Li[a], Wei Chen[a], Muhammad Khalid Khan Niazi[a]

a Department of Pathology, College of Medicine, The Ohio State University Wexner Medical Center, Columbus, OH, USA

* Corresponding author: ziyu.su@osumc.edu (Z.S.)

† *Contributed equally to this work*

## Abstract

### Background

Immunohistochemistry (IHC) enhances hematoxylin and eosin (H&E) staining by preserving tissue structure while revealing protein localization and expression, which is critical for accurate pathological diagnosis. However, IHC is costly, time-consuming, and dependent on specialized reagents, instruments, and expertise. These limitations restrict its scalability and application, especially in resource-limited settings. Computational stain translation has emerged as a promising alternative, allowing generation of IHC-equivalent images from H&E-stained slides to extract protein-level information efficiently. Yet, most existing translation models combine multiple loss terms through simple weighted summation, ignoring their interdependence and often producing images that lack optimal structural fidelity, color accuracy, or cellular detail.

### Methods

We developed a progressive network architecture that decouples the generation of color and cellular boundaries into distinct optimization stages, enabling targeted refinement of each visual component. The model was built upon an Adaptive Supervised PatchNCE framework as the baseline. To further enhance performance, we incorporated two additional loss terms: one based on 3,3′-diaminobenzidine (DAB) chromogen concentration to improve color fidelity, and another derived from image gradients to enhance cell boundary clarity.

### Results

The proposed progressive structure-color-boundary generation approach was evaluated on datasets of Human Epidermal Growth Factor Receptor 2 (HER2) and Estrogen Receptor (ER). Compared with the baseline, our method produced IHC-equivalent images with superior color consistency, clearer cellular boundaries, and improved overall structural realism. Quantitative and qualitative assessments confirmed that the proposed approach achieved substantial gains in visual quality and feature preservation.

### Conclusions

This study introduces a novel progressive framework for virtual IHC generation that effectively separates and optimizes color and structural features. The results demonstrate its potential to improve digital pathology workflows by reducing dependence on costly IHC procedures while maintaining diagnostic image quality.

# 1. Introduction

Hematoxylin and eosin (H&E) staining is effective with a variety of fixatives and provides a broad view of the cytoplasm, nucleus, and extracellular matrix, making it one of the most widely used staining methods for nearly a century [1]. H&E staining remains fundamental to histopathological evaluation; however, it lacks molecular specificity and is insufficient for distinguishing certain cell types or assessing protein expression at the biomarker level. In contrast, immunohistochemistry (IHC) enables the selective detection of antigenic proteins within tissue sections, thereby facilitating more precise characterization of cellular phenotypes and enhancing diagnostic accuracy [2]. Despite its superior ability to localize and quantify protein expression, IHC is associated with higher costs, increased tissue requirements, longer turnaround times, and the need for specialized technical expertise. These limitations hinder its routine implementation in clinical and resource-constrained settings, especially when compared to the more accessible H&E staining [3]. Consequently, the development of computational models capable of accurately translating H&E staining into IHC-equivalent representations holds promise for reducing diagnostic costs, conserving tissue samples, and minimizing the demand for specialized labor. Such advancements could significantly enhance the efficiency and scalability of histopathological workflows in both clinical and research settings.

Given the limitations of conventional staining techniques, researchers have increasingly explored deep learning-based image-to-image translation (I2IT) algorithms to computationally convert H&E-stained tissue sections into IHC-like representations. These methods, primarily based on generative adversarial networks (GANs) [4–7] and diffusion models [8, 9], have introduced notable innovations; however, none have successfully integrated the strengths of multiple methods to achieve balanced improvements across key performance metrics. For instance, enhancing adversarial loss in GANs can improve color realism but often compromises structural fidelity. This trade-off arises from the competitive nature of the objective functions within the loss architecture, which tend to optimize in conflicting directions. Linear combinations of loss components struggle to harmonize their effects, while nonlinear formulations lack generalizability, limiting the overall effectiveness of style transfer. Models employing pyramid architectures capture macroscopic tissue structures well but fall short in cellular-level detail. PSPStain has demonstrated progress in color representation, yet suffers from inconsistency [7]. The Adaptive Supervised PatchNCE (ASP) method introduces attention mechanisms to achieve more balanced improvements in both structure and coloration; however, its attention focus remains insufficiently precise, making it difficult to highlight diagnostically relevant features for clinical interpretation [10].

To overcome the limitations of existing style transfer models, we propose a novel progressive generative framework that decomposes the conversion of H&E-stained images to IHC-like representations into three sequential and functionally distinct subtasks: structure generation, color enhancement, and cell boundary refinement. Each component of the loss function is categorized according to its functional role, enabling hierarchical training of the network. In this framework, each sub-network receives the output of the preceding stage as input, and its parameters are frozen post-training to preserve feature stability and prevent interference during subsequent optimization. For the structure generation stage, we adopt the ASP model as the backbone due to its robust and balanced performance across structural and chromatic metrics. For color enhancement, we introduce the Diaminobenzidine (DAB) Enhancer module, which leverages the intensity of the 3,3'-DAB channel as the primary discriminative feature. This design choice reflects the central objective of IHC staining, which is to visualize specific protein expression, where DAB intensity serves as a proxy for the level of target protein expression. Finally, for cell boundary refinement, we propose the CellBorder_Refiner module, which utilizes high-DAB-expression regions as masks and integrates image gradient information to accurately delineate cellular boundaries, as illustrated in Figure 1.

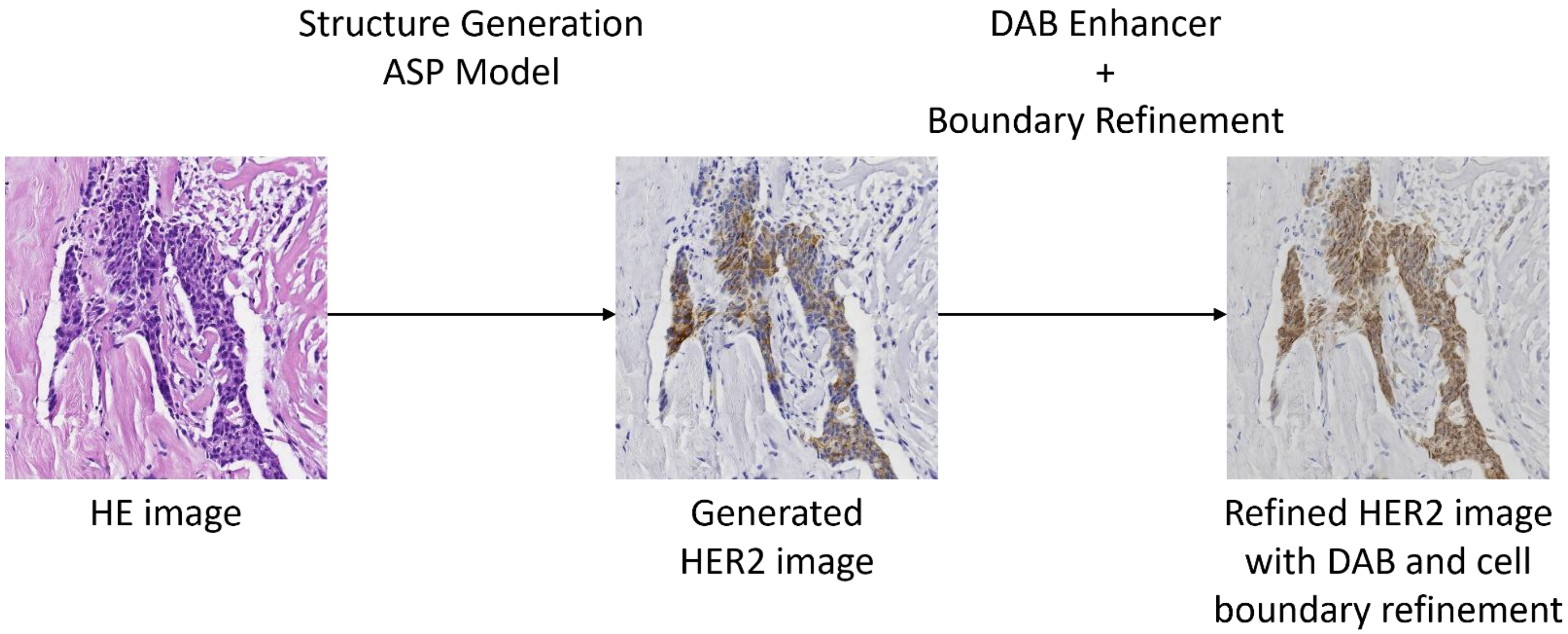

Figure 1. *A progressive generative framework for the proposed H&E to IHC image conversion. Our model will first generate the basic structure of the HER2 image from the HE image, and then refine the staining using the proposed DAB Enhancer and Boundary Refinement modules.*

The main contributions of this work are:

1. We introduce a novel, hierarchically trained generative framework comprising three sequential modules-structure generation, color enhancement, and cell boundary refinement. Each stage builds upon the output of the previous one, enabling synergistic improvements in image fidelity and interpretability.
2. We propose a targeted color enhancement strategy that leverages the 3,3' DAB channel intensity as a discriminative feature. By aligning pixel-level DAB values between real and generated images, the model more accurately conveys protein expression levels, enhancing diagnostic utility.
3. Our framework achieves substantial improvements across all quantitative metrics for the H&E-to-IHC style transfer task, demonstrating its effectiveness in generating high-fidelity, diagnostically relevant virtual IHC images.

## 2. Related Work

Loss function design is central to the success of style transfer models in histopathology, particularly for converting H&E stained images into IHC-like representations. This transformation enhances the visualization of protein expression, thereby improving diagnostic accuracy. ASP loss [5] builds upon the Supervised PatchNCE (SP) loss [11], which has demonstrated strong performance in H&E to IHC style transfer tasks. ASP introduces enhancements inspired by the robustness of contrastive learning [12] to label noise, aiming to reduce the detrimental effects of inconsistent patch positioning during training. While ASP provides a strong foundation, its integration with other loss functions remains limited in scope. Despite its advantages, current models typically employ a linear combination of ASP with adversarial loss [13], PatchNCE loss [6], and Gaussian pyramid reconstruction loss [14]—all of which primarily emphasize structural preservation. This approach yields images with coherent tissue architecture but fails to adequately incorporate chromatic and boundary-specific objectives, often compromising the accuracy of protein expression levels and cellular delineation.

To overcome these limitations and build upon Contribution #1, we extend the ASP-based framework by introducing a modular training pipeline—Structure-to-Color-to-Cell Border Network—that enables independent optimization of each visual component. This architecture preserves the robustness of ASP while achieving a more balanced integration of structural, chromatic, and boundary-level features. In addition to structural and adversarial losses, chromatic fidelity is critical for accurate IHC representation—a challenge we address through targeted color deconvolution. Color deconvolution is a widely adopted technique in histopathological image analysis, particularly for separating the contributions of individual stains in H&E and IHC images [15]. It enables the decomposition of mixed color signals into distinct staining channels, such as isolating the 3,3'-DAB channel, which is critical for quantifying protein expression in IHC. In our framework (Contribution #2), color deconvolution is employed during the color enhancement stage to extract pixel-level DAB concentrations from real IHC images used in training. These values serve as a reference for computing a DAB-based loss, guiding the model to generate synthetic IHC images with more accurate protein localization and expression intensity. Unlike prior approaches that rely on generic chromatic losses, our DAB-based loss directly aligns stain intensity with biological targets, enhancing diagnostic utility. By integrating structure-aware, stain-specific, and boundary-refining losses into a unified framework, our approach addresses key limitations in existing H&E-to-IHC translation models and supports the substantial performance gains outlined in Contribution #3.

# 3. Method

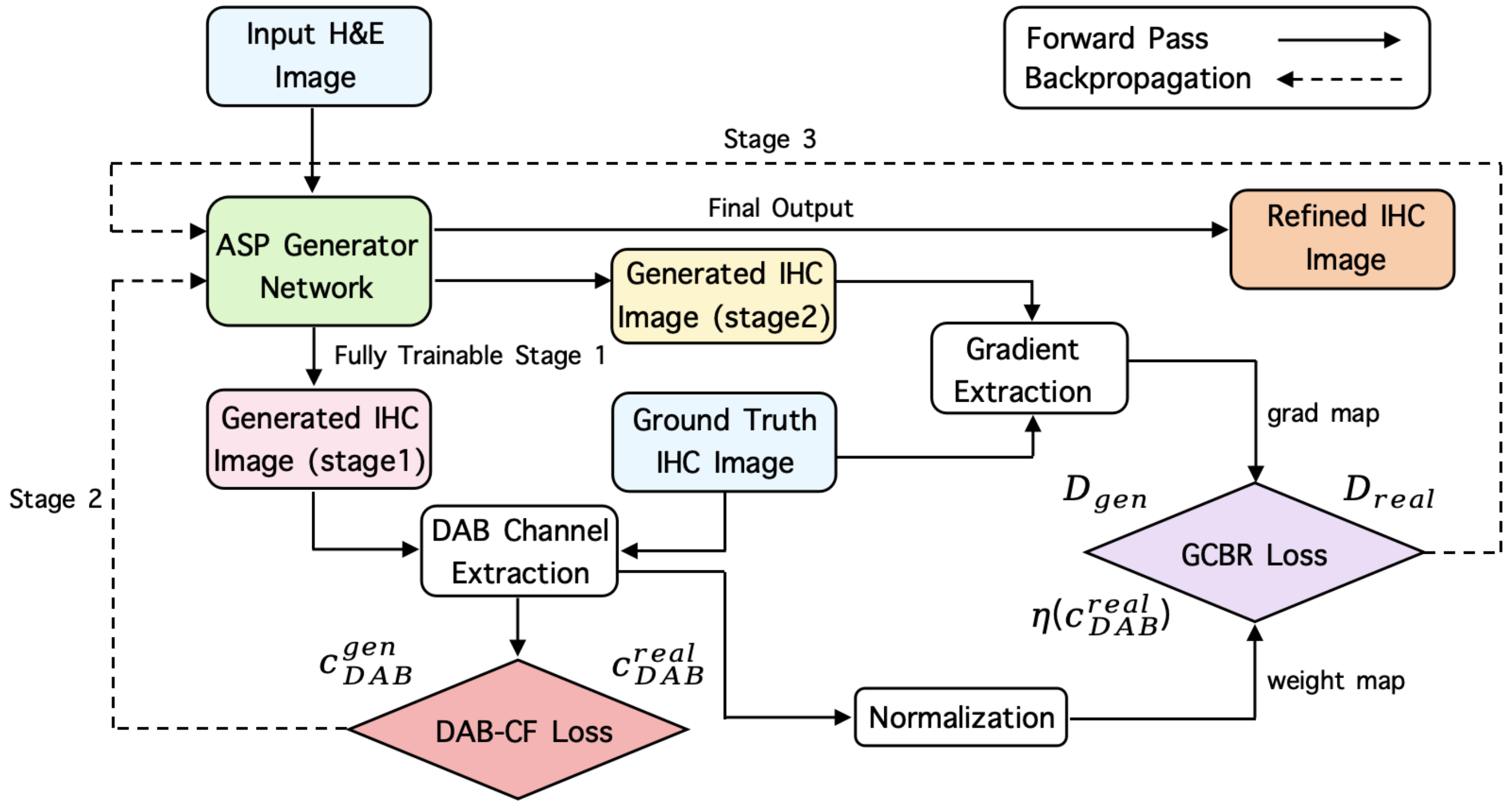

**Figure 2.** *Overview of ProgASP. The ASP Generator is trained in three stages: Stage 1 ensures structural alignment, Stage 2 enforces DAB-guided color fidelity, and Stage 3 refines cell boundaries via gradient-guided supervision, yielding the final Refined IHC Image.*

We propose ProgASP, a progressive generative framework for synthesizing IHC-like representations from H&E-

stained images. The framework consists of three sequential stages: (1) structural generation, (2) color fidelity enhancement, and (3) cell boundary refinement. Each stage builds upon the output of the previous one, enabling synergistic improvements in image fidelity and interpretability. The overview of ProgASP is illustrated in Figure 2. In Stage 1, the ASP Generator (encoder and decoder) is trained end-to-end to synthesize an initial IHC image. In Stage 2, the encoder is frozen and only the decoder is updated using DAB-guided supervision to improve chromatic fidelity. In Stage 3, the decoder is further refined with gradient-guided loss to sharpen cellular boundaries, producing the final refined IHC image.

## 3.1 Adaptive Supervised PatchNCE (ASP) Loss

Since H&E and IHC images are typically obtained from adjacent but distinct tissue sections, pixel-level misalignment is a common and unavoidable issue in histopathology image translation tasks. Therefore, we choose the ASP model based on the ASP loss function to perform the structural transformation of the initial H&E image to IHC staining in our framework [5]. This essentially enhances the robustness of the generated IHC image's structure. The ASP model can achieve this through a properly designed ASP loss.

Mathematically, the SP loss at a given patch location is defined as an InfoNCE-style contrastive loss:

$$\mathcal{L}_{SP} = -\log \frac{\exp(z_{\hat{y}} \cdot z_y / \tau)}{\exp(z_{\hat{y}} \cdot z_y / \tau) + \sum_{n=1}^{N} \exp(z_{\hat{y}} \cdot z_n^- / \tau)},$$

where $z_{\hat{y}}$ is the embedding of the generated IHC patch, $z_y$ is the embedding of the corresponding patch from the ground-truth IHC image, and $z_n^-$ are embeddings of negative (non-matching) patches. The temperature parameter $\tau$ controls the contrast.

ASP further enhances its ability to manage ground-truth inconsistencies through an adaptive weighting mechanism. During training, it dynamically adjusts the contribution of each patch based on the similarity between the generated embedding and the ground-truth embedding, assigning higher weights to patches with greater similarity. This stabilizes the structural features of the generated image. Mathematically, it can be expressed as:

$$w_t(z_{\hat{y}}, z_y) = (1 - \mathrm{g}(t/T)) \cdot 1.0 + \mathrm{g}(t/T) \cdot \mathrm{h}(z_{\hat{y}} \cdot z_y),$$

where g(t/T) is a scheduling function that gradually increases the influence of the weighting over training iterations, and h(·) is a monotonic function mapping cosine similarity to a weight value.

The final ASP loss aggregates patch-level contrastive objectives across multiple feature map layers $l$ of the backbone network (e.g., encoder layers), where each layer provides embeddings at a distinct semantic scale. Formally:

$$\mathcal{L}_{ASP} = E_{(x,y)\sim(X,Y)} \sum_{l=1}^{L} \sum_{s=1}^{S_l} \frac{w_t(z_{s,\hat{y}}^l, z_{s,y}^l)}{W_t^l} \cdot \mathcal{L}_{InfoNCE}(z_{s,\hat{y}}^l, z_{s,y}^l, \{z_{s,n}^{-l}\}_{n=1}^N),$$

where $z_{s,\hat{y}}^l$ and $z_{s,y}^l$ denote the embeddings of generated and real IHC patches at spatial location $s$ in layer $l$, and $\{z_{s,n}^{-l}\}_{n=1}^N$ are negative samples from non-matching patches. The weight term $w_t(\cdot)$ adaptively balances patch contributions, and $W_t^l$ normalizes across scales to stabilize optimization.

The loss function used in the first stage (structure generation) is formulated as:

$$\mathcal{L}_{Struct} = \mathcal{L}_{adv} + \lambda_{PatchNCE}\mathcal{L}_{PatchNCE} + \lambda_{ASP}\mathcal{L}_{ASP} + \lambda_{GP}\mathcal{L}_{GP},$$

where $\mathcal{L}_{adv}$ is the adversarial loss [13], $\mathcal{L}_{PatchNCE}$ is the unsupervised PatchNCE loss [11] and $\mathcal{L}_{GP}$ is the Gaussian Pyramid loss [14].

In summary, we employ the ASP loss formulation [5] as the core objective of the structure generation stage. By combining ASP with adversarial and Gaussian Pyramid losses within our architecture, this stage achieves structurally consistent IHC synthesis even under pixel-level misalignment, thereby stabilizing tissue morphology and laying the groundwork for subsequent stages. However, this stage remains focused exclusively on structural fidelity and does not impose constraints on staining quality. Consequently, the generated IHC images may still suffer from incomplete staining, inaccurate staining intensity, and blurred or anatomically imprecise cellular boundaries. These shortcomings motivate the introduction of additional objectives that explicitly enforce color fidelity and cell boundary clarity, which are detailed in Sections 3.2 and 3.3.

## 3.2. DAB-Guided Color Fidelity (DAB-CF)

To improve chromatic fidelity, we introduce the DAB-Guided Color Fidelity (DAB-CF) module. This loss leverages the biologically meaningful 3,3′-DAB channel to directly constrain the chromatic and biochemical fidelity of synthesized IHC images. Since the DAB signal reflects true protein expression, enforcing consistency ensures that generated IHC images are not only visually realistic but also diagnostically reliable.

1. Conversion to Optical Density (OD) Space

Let an RGB histopathology image be denoted as $I \in R^{H\times W\times 3}$, normalized to $[0, 1]$. We first convert the pixel intensity at location $(x, y)$ into optical density ($OD$) space, which linearizes stain absorbance:

$$OD(x, y) = -\log_{10}\left(\frac{I(x, y) + \epsilon}{I_0}\right),$$

where $I_0$ is the reference white intensity and $\epsilon$ is a small constant to ensure numerical stability.

2. Stain Separation

Following Ruifrok and Johnston's framework [15], we construct a stain matrix $M \in \mathbb{R}^{3\times 3}$, where each row corresponds to the normalized absorbance vector of hematoxylin, eosin, and DAB stains in RGB space. For each pixel, the concentration vector is obtained by:

$$c(x,y) = M^{-1} \cdot OD(x, y),$$

where $c(x,y) = [c_H(x,y), c_E(x,y), c_{DAB}(x,y)]^T$. The DAB concentration $c_{DAB}(x,y)$ represents the localized DAB concentration and directly reflects target protein expression.

3. DAB-CF Objective

To enforce pixel-level agreement between generated and real DAB channels, we define the DAB-CF as the mean squared error (MSE) between the extracted DAB channel of the real IHC image $c_{DAB}^{real}$ and that of the generated image $c_{DAB}^{gen}$:

$$\mathcal{L}_{DAB-CF} = \frac{1}{N}\sum_{x,y}(c_{DAB}^{real}(x,y) - c_{DAB}^{gen}(x,y))^2,$$

where $N$ is the total number of pixels. This loss enforces pixel-level agreement in DAB intensity, improving protein localization and expression fidelity.

### 3.3 Gradient-guided Cell Boundary Refinement (GCBR)

Although the ASP and DAB-CF objectives ensure structural alignment and chromatic fidelity, the generated IHC images may still suffer from blurred or anatomically imprecise cellular boundaries, particularly for membrane-bound protein markers such as HER2. Accurate delineation of cell borders is critical for pathological interpretation, as membrane-localized proteins often serve as biomarkers for diagnostic and therapeutic decision-making. To address this limitation, we propose the Gradient-Guided Cell Boundary Refinement (GCBR) module, which explicitly enforces high-frequency consistency along cellular boundaries by combining image gradient supervision with DAB-based spatial localization.

1. Gradient Extraction

Let $\nabla(\cdot)$ denote the grayscale intensity of an IHC image at pixel $(x, y)$, The local structural variation is quantified via the image gradient:

$$\nabla I(x,y) = \begin{bmatrix} \frac{\partial I}{\partial x} & \frac{\partial I}{\partial y} \end{bmatrix}, \ \|\nabla I(x,y)\| = \sqrt{(\frac{\partial I}{\partial x})^2 + (\frac{\partial I}{\partial y})^2}.$$

In practice, we approximate $\nabla I$ using the Sobel operator [17], which computes first-order derivatives along the horizontal and vertical directions. For each pixel, we obtain the gradient magnitude maps of the generated and real IHC images:

$$D_{gen}(x,y) = \|\nabla I_{gen}(x,y)\|, \ D_{real}(x,y) = \|\nabla I_{real}(x,y)\|.$$

2. Weighted Gradient Consistency Loss

To ensure refinement is concentrated in biologically meaningful regions, we incorporate information from the DAB channel (Section 3.2). A soft weighting map $\eta(x, y)$ is constructed by normalizing the real DAB concentration:

To prioritize diagnostically meaningful regions, we introduce a soft weight derived from the normalized DAB intensity:

$$\eta(x,y) = norm\left(c_{DAB}^{real}(x,y)\right), \ \eta(x,y) \in [0,1],$$

where $c_{DAB}^{real}(x, y)$ is extracted via color deconvolution [15]. This weighting emphasizes protein-rich zones near membranes, avoiding spurious supervision in irrelevant background regions.

The GCBR loss is then defined as:

$$\mathcal{L}_{GCBR} = \frac{1}{\sum_{x,y} \eta(x,y)} \sum_{x=1}^{H} \sum_{y=1}^{W} \eta(x,y)(D_{gen}(x,y) - D_{real}(x,y))^2.$$

This formulation generalizes the hard mask approach [17] by adopting a continuous weighting scheme, which stabilizes training and provides finer control over gradient supervision.

By complementing ASP with DAB-CF and GCBR, our ProgASP overcomes two critical shortcomings of existing ASP-based models: (1) the absence of explicit chromatic constraints and (2) the lack of boundary-focused refinement. While ASP ensures robust structural alignment, DAB-CF enforces biologically meaningful color fidelity, and GCBR sharpens cellular boundaries through gradient-guided, DAB-weighted supervision. This multi-objective integration results in IHC image synthesis that not only preserves tissue architecture but also exhibits faithful staining patterns and anatomically precise cell morphology—factors essential for downstream pathological assessment.

**Overall Objective.** We adopt a progressive, stage-wise optimization strategy rather than joint training. The total objective at each stage is defined as:

$$\mathcal{L}_{total}(stage) = \begin{cases} \mathcal{L}_{Struct}, & stage\ 1 \\ \lambda_{DAB}\ \mathcal{L}_{DAB-CF}, & stage\ 2 \\ \lambda_{grad}\mathcal{L}_{GCBR}, & stage\ 3 \end{cases}$$

At each stage, only the corresponding subnetwork is updated, while the parameters of all previously trained modules remain frozen. Specifically, at stage 2, we train an independent ASP generator network using the output images from stage 1 as inputs. Then, at stage 3, we train another ASP generator network using the output from stage 2 as inputs.

This progressive formulation ensures stable structural alignment in the first stage, chromatic fidelity in the second, and accurate boundary refinement in the final stage.

# 4. Experiments

**Dataset.** In this work, we employed the Multi-domain Immunohistochemical Stain Translation (MIST) dataset [5], which provides paired histological images of H&E and IHC staining across multiple biomarkers. We concentrated on two subsets of clinical relevance to breast cancer: HER2 (human epidermal growth factor receptor 2) and ER (estrogen receptor). The HER2 receptor is localized to the cell membrane, whereas ER is found in the nucleus. For the HER2 subset, 4642 non-overlapping paired patches (1024 × 1024 pixels) for training and 1000 for testing were extracted from 64 whole-slide images (WSIs). For the ER subset, 4153 training pairs and 1000 testing pairs were collected from 56 WSIs. Each H&E patch is paired with an IHC counterpart from an adjacent section, leading to imperfect pixel-level alignment but consistent tissue morphology, which supports supervised translation tasks.

**Implementation details.** Our training process strictly sequentially steps through structure generation, color enhancement, and cell boundary refinement. At each stage, only the corresponding subnetwork is updated, while the parameters of the previous training stage are frozen. Following [5], we set the weighting coefficients for the structure stage as $\lambda_{PatchNCE}$ = 10.0, $\lambda_{ASP}$ = 10.0, and $\lambda_{GP}$ = 10.0. For the subsequent stages, we set $\lambda_{DAB}$ = 0.5 during color enhancement and $\lambda_{grad}$ = 1.0 during cell boundary refinement. These coefficients act as scaling factors, balancing the contributions of their respective objectives. All models were implemented in PyTorch and trained on a single NVIDIA A100 GPU. The entire progressive training process spanned three phases and took approximately 12 hours to complete. During the training, we use the Adam optimizer [19] with a batch size of 40 and an initial learning rate of $2 \times 10^{-4}$.

**Experimental Setup.** For both HER2 and ER subsets of the MIST dataset, we follow the official splits provided in [5], using the designated training and test sets (HER2: 4642/1000; ER: 4153/1000). Within the training set, 10% of the patches are randomly held out as a validation set for monitoring convergence and early stopping. All quantitative results reported in Table 1 are obtained from the held-out **test set**, ensuring an unbiased comparison across methods.

Our evaluation methods include paired and unpaired evaluations. For paired assessment, we use the Structural Similarity Index (SSIM) [20], Peak Signal-to-Noise Ratio (PSNR) [21], and Perceptual Hash Value (PHV) [22] to measure the goodness of fit, and the Gradient Mean Squared Error (Gradient MSE) [23] to quantify the sharpness of boundaries. For unpaired evaluation, we use the Fréchet Inception Distance (FID) [24] and Kernel Inception Distance (KID) to assess overall distribution consistency.

**Quantitative Results.** Table 1 presents the comprehensive comparison results of the baseline ASP model, the Stable Diffusion model [25], and our proposed ProgASP framework. After applying our restructured stage-wise optimization, ProgASP consistently outperforms existing methods across nearly all evaluation metrics. On the HER2 dataset, ProgASP surpasses both Stable Diffusion and ASP in every metric. In particular, under the PHV metric, ProgASP achieves the lowest values across all layers, indicating reduced perceptual discrepancy with the ground truth. It also achieves the best unpaired metrics, lowering the FID to 49.6 (14.2% below ASP) and the KID to 0.0167 (7% below ASP), while also obtaining the lowest gradient MSE (0.002234). On the ER dataset, ProgASP demonstrates similar advantages, achieving an SSIM of 0.2034 and a PSNR of 13.8147, both state-of-the-art. It further yields the best PHV results across all layers, with the lowest FID (40.1) and KID (0.0062), outperforming both ASP and Stable Diffusion.

Table 1. *Different methods were quantitatively compared on the HER2 and ER subsets of the MIST dataset. The best values are highlighted.*

| Datasets | Methods | SSIM↑ | PSNR↑ | PHV↓ (layer1) | PHV↓ (layer2) | PHV↓ (layer3) | PHV↓ (layer4) | FID↓ | KID↓ | Gradient MSE↓ |
|---|---|---|---|---|---|---|---|---|---|---|
| HER2 | Stable-Diffusion | 0.1220 | 10.7315 | 0.7007 | 0.6779 | 0.5019 | 0.9070 | 307.5 | 0.2399 | 0.002245 |
| | ASP | 0.1868 | 13.8197 | 0.4822 | 0.4539 | 0.2724 | 0.8225 | 57.8 | 0.0187 | 0.002236 |
| | ProgASP (ours) | **0.2138** | **14.7023** | **0.4739** | **0.4379** | **0.2631** | **0.8166** | **49.6** | **0.0167** | **0.002234** |
| ER | Stable-Diffusion | 0.1338 | 10.4536 | 0.7202 | 0.6915 | 0.5191 | 0.9138 | 318.3 | 0.2454 | 0.002486 |
| | ASP | 0.2006 | 13.6628 | 0.4729 | 0.4211 | 0.2775 | 0.8261 | 43.7 | 0.0099 | **0.002468** |
| | ProgASP (ours) | **0.2034** | **13.8147** | **0.4563** | **0.4178** | **0.2739** | **0.8244** | **40.1** | **0.0062** | 0.002470 |

**Qualitative Results.** As shown in Figure 3, our framework can generate more realistic and stable IHC images with a superior visual experience in terms of structure, color, and cell boundaries, and is consistent with real images. This demonstrates that our dedicated enhancement module has been successful and can provide better diagnostic assistance to pathologists in subsequent applications.

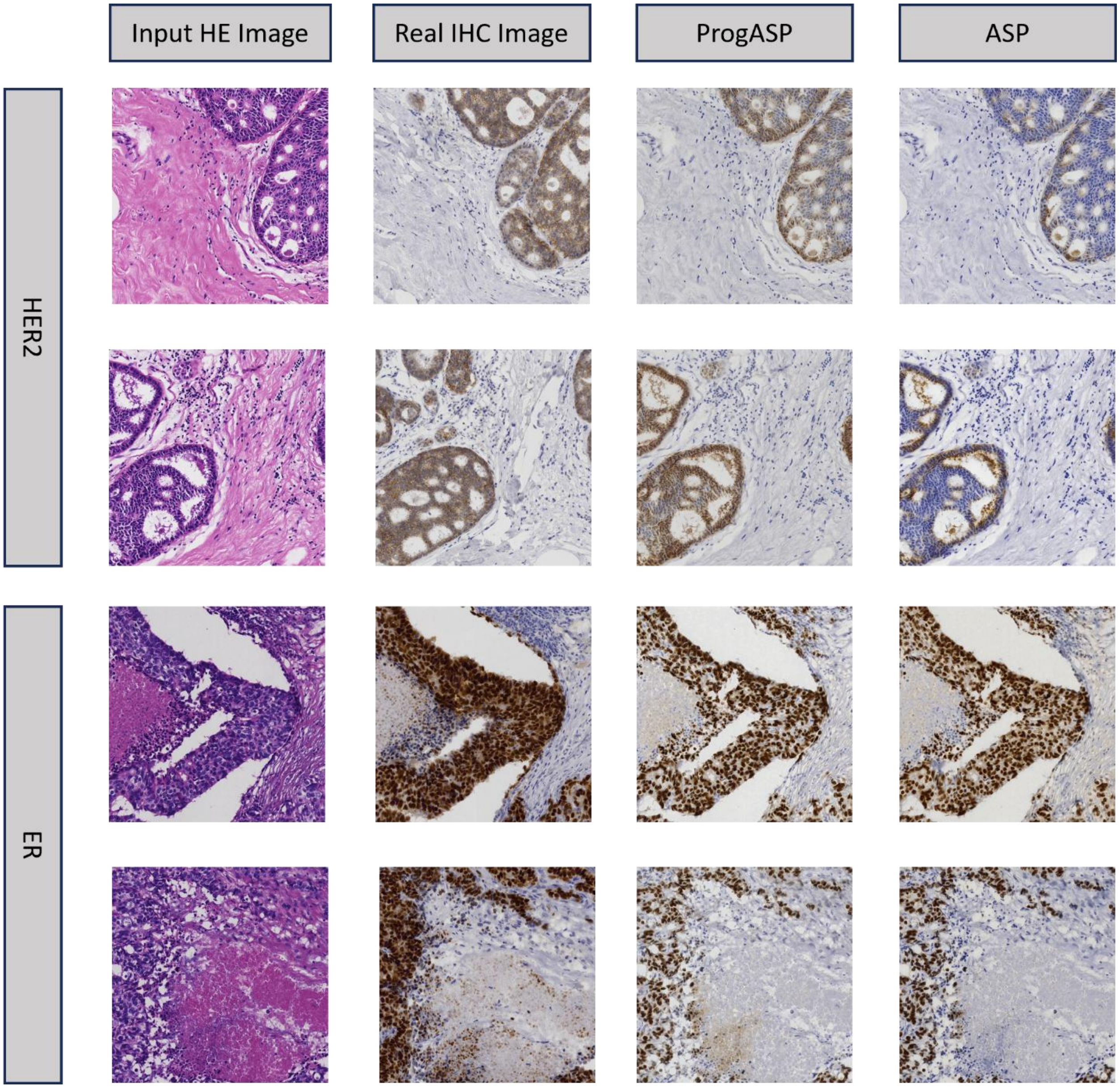

Figure 3. *Qualitative comparison on the MIST dataset (HER2 and ER subsets). Each triplet shows the input H&E image, the corresponding ground-truth IHC, and the IHC generated by ProgASP. The generated results reproduce staining intensity and structural details with high chromatic fidelity and morphological consistency.*

## 5. Discussion

The results of our study demonstrate that integrating the ASP loss with biologically and morphologically guided enhancements significantly improves the fidelity of H&E-to-IHC image translation. Our framework addresses a critical gap in existing generative models: the lack of explicit constraints on staining quality and cellular morphology. By introducing the DAB-Guided Color Fidelity and Gradient-Guided Cell Boundary Refinement modules, we move beyond structural alignment to enforce biochemical realism and diagnostic clarity.

Compared to prior approaches such as Stable Diffusion and the original ASP model, our method consistently outperforms across both paired and unpaired evaluation metrics. The improvements in SSIM, PSNR, and PHV indicate better structural and perceptual alignment with ground-truth IHC images, while reductions in FID and Gradient MSE reflect enhanced distributional consistency and sharper morphological boundaries. These gains are especially notable in clinically relevant biomarkers like HER2 and ER, where accurate localization and intensity of protein expression are essential for diagnostic decision-making.

Importantly, our progressive training strategy—where each module is trained sequentially with frozen parameters from previous stages—ensures stability and modularity. This design allows for targeted optimization of structure, color, and boundary features without interference between objectives. It also opens the door for future extensions, such as incorporating additional stain types or integrating domain-specific priors (e.g., cell type annotations or spatial transcriptomics data).

Despite these strengths, our framework has limitations. First, while the DAB-CF module enforces chromatic fidelity, it relies on stain separation techniques that may be sensitive to variations in staining protocols and scanner calibration. Second, the GCBR module assumes that gradient magnitude correlates with boundary clarity, which may not hold in all tissue types or pathological conditions. Third, our evaluation is limited to the MIST dataset; broader validation across diverse tissue types, staining protocols, and clinical settings is necessary to confirm generalizability.

## 6. Conclusion

We present a progressive generative framework for translating H&E-stained histopathology images into diagnostically meaningful IHC images. By integrating the ASP loss with two novel modules—DAB-Guided Color Fidelity and Gradient-Guided Cell Boundary Refinement—our approach addresses key limitations in existing methods, including structural misalignment, incomplete staining, and poor boundary localization. Our experiments on the MIST dataset, particularly for HER2 and ER biomarkers, demonstrate that the proposed framework significantly outperforms both Stable Diffusion and the original ASP model across structural, perceptual, and distributional metrics. These improvements validate the effectiveness of our multi-stage training strategy and biologically informed loss functions. Beyond quantitative performance, our method offers practical value for surgical pathology by generating IHC images that are not only visually realistic but also biochemically and morphologically faithful. This capability is especially valuable in settings where IHC staining is unavailable, delayed, or cost-prohibitive, opening new possibilities for AI-assisted diagnostics.

Future work will focus on expanding the framework to additional biomarkers, validating across diverse datasets, and optimizing for real-time deployment in clinical workflows. This may include model pruning and the integration of explainable AI techniques to enhance interpretability and speed. Ultimately, our approach contributes to the advancement of computational pathology by bridging the gap between image synthesis and diagnostic utility, laying the groundwork for clinically relevant AI-driven histopathology.

## Ethics approval and consent to participate

Not applicable.

## Consent for publication

We have received the consent for publication from all authors

## Availability of data and material

Our dataset is available through https://bupt-ai-cz.github.io/BCI/

## Competing interest

Not applicable

## Authors' contributions

Y.K. performed data preprocessing, designed and conducted experiments, validation, and contributed to manuscript writing. Z.S. performed data preprocessing, designed and conducted experiments, validation, and contributed to manuscript writing. T.W. performed data preprocessing, designed and conducted experiments, validation, and contributed to manuscript writing. M.K.K.N. provided feedback on the study design and validation, edited the manuscript, conceptualized and designed the study, and supervised the research. Z.L. provided clinical assessment and contributed to manuscript editing. W.C. provided clinical assessment and contributed to manuscript editing. Z.S. also conceptualized and designed the study, and supervised the research.

## Clinical Trial Number

Not applicable

## Funding declaration

This work was supported in part by the National Cancer Institute under grant R01 CA276301 (PIs: Niazi and Chen), the National Institute on Deafness and Other Communication Disorders under grant R01 DC020715 (PIs: Moberly and Gurcan), and the National Heart, Lung, and Blood Institute under grant R01 HL177046-01 (PI: Hachem). Additional support was provided by Pelotonia under IRP CC13702 (PIs: Niazi, Vilgelm, and Roy), and by The Ohio State University Department of Pathology and Comprehensive Cancer Center.

## Acknowledgement

The authors thank all funding agencies for their generous support. The content is solely the responsibility of the authors and does not necessarily represent the official views of the National Cancer Institute, the National Institutes of Health, Pelotonia, or The Ohio State University.